\documentclass[10pt,twocolumn,letterpaper]{article}

\usepackage[pagenumbers]{cvpr} 

\usepackage{graphicx}
\usepackage{amsmath}
\usepackage{stix}
\usepackage{booktabs}
\usepackage{threeparttable}
\usepackage{xspace}
\usepackage{graphicx}
\usepackage{color}
\usepackage[misc]{ifsym}
\usepackage[normalem]{ulem}
\usepackage{multirow}
\usepackage{float}
\usepackage{amsfonts}
\usepackage{bm}
\usepackage{array}
\usepackage[table]{xcolor}
\usepackage{colortbl}
\usepackage{diagbox}
\usepackage{rotating}
\usepackage{booktabs}
\usepackage{overpic}
\usepackage{enumitem}
\usepackage{colortbl}
\usepackage{algorithm}
\usepackage{algorithmic}
\usepackage{makecell}
\usepackage{flushend}
\usepackage{ifsym}
\usepackage[numbers,sort]{natbib}
\definecolor{myorange}{RGB}{255,100,3}
\definecolor{mygray}{gray}{.85}
\definecolor{mygray1}{gray}{.7}
\definecolor{mygray2}{gray}{.93}
\definecolor{mygray3}{gray}{.90}

\newcommand{\numvideo}{2,149\xspace}
\newcommand{\numobject}{5,200\xspace}
\newcommand{\numclass}{36\xspace}
\newcommand{\nummask}{431,725\xspace}
\newcommand{\ourdataset}{MOSE\xspace}

\makeatletter
\newcommand{\thickhline}{%
    \noalign {\ifnum 0=`}\fi \hrule height 1pt
    \futurelet \reserved@a \@xhline
}
\makeatother

\makeatletter
\DeclareRobustCommand\onedot{\futurelet\@let@token\@onedot}
\def\@onedot{\ifx\@let@token.\else.\null\fi\xspace}

\def\eg{\emph{e.g}\onedot} 
\def\ie{\emph{i.e}\onedot} 
 
\def\etc{\emph{etc}\onedot} \def\vs{\emph{vs}\onedot}
 
\def\etal{\emph{et al}\onedot}

\makeatother

\usepackage[pagebackref,breaklinks,colorlinks]{hyperref}

\usepackage[capitalize]{cleveref}
\crefname{section}{Sec.}{Secs.}
\Crefname{section}{Section}{Sections}
\Crefname{table}{Table}{Tables}
\crefname{table}{Tab.}{Tabs.}

\def\cvprPaperID{*****} 
\def\confName{CVPR}
\def\confYear{2023}

\begin{document}

\title{MOSE: A New Dataset for Video Object Segmentation in Complex Scenes}

\author{
Henghui Ding$^1$
\quad
Chang Liu$^1$
\quad
Shuting He$^2$
\quad
Xudong Jiang$^1$
\quad
Philip H.S. Torr$^3$
\quad
Song Bai$^4$\\
$^1$Nanyang Technological University
\quad
$^2$Zhejiang University
\quad
$^3$University of Oxford
\quad
$^4$ByteDance\\
\vspace{-3mm}
\\
\href{https://henghuiding.github.io/MOSE}{https://henghuiding.github.io/\ourdataset}
}

\twocolumn[{%
\renewcommand\twocolumn[1][]{#1}%
\maketitle 
\begin{center} 
\centering 
\vspace{-3mm}
\captionsetup{type=figure}
\includegraphics[width=0.999\textwidth]{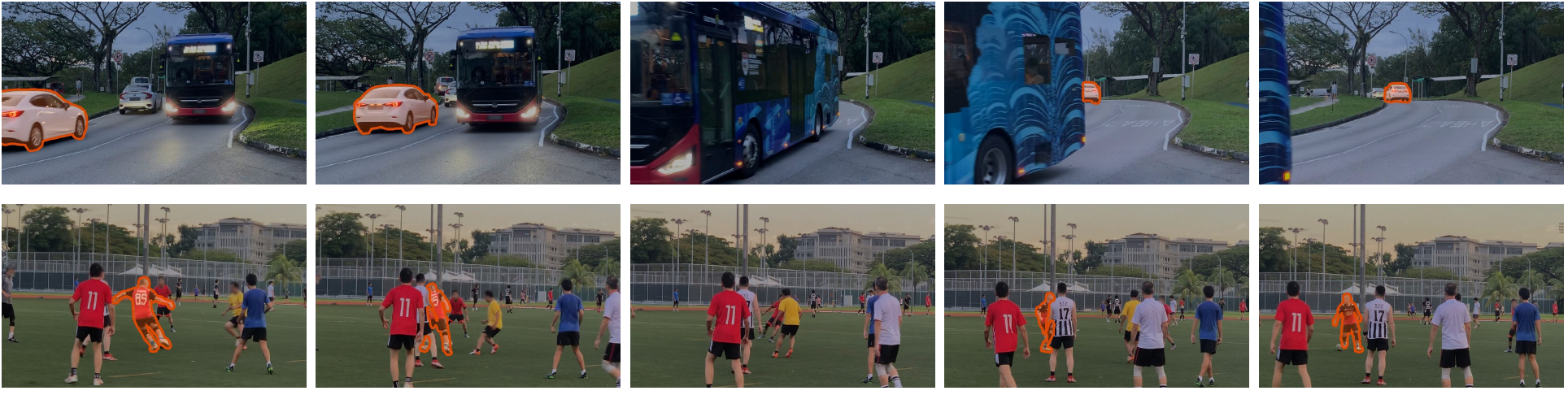} 
\vspace{-7mm}
\captionof{figure}{\small Examples of video clips from the co\textbf{M}plex video \textbf{O}bject \textbf{SE}gmentation (\ourdataset) dataset. The selected target objects are masked in \textcolor{myorange}{orange} ${\color{myorange}\mdblksquare}$. The most notable feature of MOSE is complex scenes, including the disappearance-reappearance of objects, small/inconspicuous objects, heavy occlusions, crowded environments, \etc. For example, the target player in the 2nd row turns around when reappearing in the 4th and 5th columns after disappearing in the 3rd column, bringing challenges in re-identifying him. Most videos in MOSE contain crowded and occluded objects with the target object seldom being the salient one. The goal of MOSE dataset is to provide a platform that promotes the development of more comprehensive and robust video object segmentation algorithms.
}
\vspace{3mm}
\label{Fig:teaser}
\end{center}
}]

\renewcommand{\thefootnote}{\fnsymbol{footnote}}
\footnotetext[0]{${\textrm{\Letter}}$ henghui.ding@gmail.com }
\begin{abstract}
   Video object segmentation (VOS) aims at segmenting a particular object throughout the entire video clip sequence. The state-of-the-art VOS methods have achieved excellent performance (\eg, \textbf{90+\%} $\mathcal{J}\&\mathcal{F}$) on existing datasets. However, since the target objects in these existing datasets are usually relatively salient, dominant, and isolated, VOS under complex scenes has rarely been studied. To revisit VOS and make it more applicable in the real world, we collect a new VOS dataset called co\textbf{M}plex video \textbf{O}bject \textbf{SE}gmentation (\ourdataset) to study the tracking and segmenting objects in complex environments. \ourdataset contains \numvideo video clips and \numobject objects from \numclass categories, with \nummask high-quality object segmentation masks. The most notable feature of \ourdataset dataset is complex scenes with crowded and occluded objects. The target objects in the videos are commonly occluded by others and disappear in some frames. To analyze the proposed \ourdataset dataset, we benchmark 18 existing VOS methods under 4 different settings on the proposed \ourdataset dataset and conduct comprehensive comparisons. The experiments show that current VOS algorithms cannot well perceive objects in complex scenes. For example, under the semi-supervised VOS setting, the highest $\mathcal{J}\&\mathcal{F}$ by existing state-of-the-art VOS methods is only \textbf{59.4\%} on \ourdataset, much lower than their $\sim$90\% $\mathcal{J}\&\mathcal{F}$ performance on DAVIS. The results reveal that although excellent performance has been achieved on existing benchmarks, there are unresolved challenges under complex scenes and more efforts are desired to explore these challenges in the future. The proposed \ourdataset dataset has been released at~\href{https://henghuiding.github.io/MOSE}{https://henghuiding.github.io/\ourdataset}.
\end{abstract}

\section{Introduction}
\label{sec:intro}

Video object segmentation (VOS)~\cite{davis2016,davis2017,youtube_vos} aims at segmenting a particular object, \eg, the dominant objects or the objects indicated by users, throughout the entire video sequence. It is one of the most fundamental and challenging computer vision tasks and has a crucial role in many piratical applications that involve video analysis and understanding, \eg, self-driving vehicle, augmented reality, video editing, \etc. There are many different settings for VOS, for example, semi-supervised VOS~\cite{DBLP:conf/cvpr/CaellesMPLCG17,Park_2021_CVPR} that gives the first-frame mask of the target object, unsupervised VOS~\cite{jain2017fusionseg,cheng2017segflow} that automatically finds primary objects, and interactive VOS~\cite{chen2018blazingly,oh2019fast} that relies on the user's interactions of the target object. Video object segmentation has been extensively studied in the past using traditional techniques or deep learning methods. Especially, the deep-learning-based methods have greatly improved the performance of video object segmentation and surpassed the traditional techniques by a large margin.

Current state-of-the-art methods have achieved very high performance on two of the most commonly-used VOS datasets DAVIS~\cite{davis2016, davis2017} and YouTube-VOS~\cite{youtube_vos}. For example, XMem~\cite{XMem} achieves \textbf{92.0\%} $\mathcal{J}\&\mathcal{F}$ on DAVIS~2016~\cite{davis2016}, \textbf{87.7\%} $\mathcal{J}\&\mathcal{F}$ on DAVIS~2017~\cite{davis2017}, and \textbf{86.1\%} $\mathcal{G}$ on YouTube-VOS~\cite{youtube_vos}. With such a high performance, it seems that the video object segmentation has been well resolved. However, do we really perceive objects in realistic scenarios? To answer this question, we revisit video object segmentation under realistic and more complex scenes. The target objects in existing datasets~\cite{youtube_vos,davis2016,davis2017} are usually salient and dominant. In real-world environments, isolated and salient objects rarely appear while complex and occluded scenes happen frequently.
To evaluate current state-of-the-art VOS methods under more complex scenes, we collect \numvideo videos with complex scenes and form a new large-scale challenging video object segmentation benchmark, termed co\textbf{M}plex video \textbf{O}bject \textbf{SE}gmentation (\ourdataset).
Specifically, there are \numobject objects from \numclass categories in \ourdataset, with \nummask high-quality segmentation masks. As shown in \Cref{Fig:teaser}, the most notable feature of \ourdataset is complex environments, including disappearance-reappearance of objects, small/inconspicuous objects, heavy occlusions, crowded environments, \etc. For example, the white sedan in the first row of \Cref{Fig:teaser} is occluded by the bus, and the heaviest occlusion in 3rd image makes the sedan totally disappear. In the second row of \Cref{Fig:teaser}, the target player in the crowd is inconspicuous and disappears in the third frame due to the occlusions of the crowd. When the target player reappears, he turns around and shows a different appearance from the first two frames, which makes him very difficult to be tracked. The heavy occlusion and disappearance of objects under complex scenes bring great challenges to video object segmentation. We wish to promote video object segmentation research in complex environments and make VOS applicable in the real world.

To analyze the proposed \ourdataset dataset, we retrain and evaluate some existing VOS methods on \ourdataset. Specifically, we retrain 6 state-of-the-art methods under the semi-supervised setting using mask as the first-frame reference, 2 methods under the semi-supervised setting using bounding box as the first-frame reference, 3 methods under the multi-object zero-shot video object segmentation setting, and 7 methods under interactive setting. The experimental results show that videos of complex scenes make the current state-of-the-art VOS methods less pronounced, especially in terms of tracking objects that disappear for a while due to occlusions. For example, the $\mathcal{J}\&\mathcal{F}$ performance of XMem~\cite{XMem} on DAVIS 2016 is \textbf{92.0\%} but drop to \textbf{57.6\%} on \ourdataset, the $\mathcal{J}\&\mathcal{F}$ performance of DeAOT~\cite{DeAOT} on DAVIS 2016 is \textbf{92.9\%} but drop to \textbf{59.4\%} on \ourdataset, which consistently reveal the difficulties brought by complex scenes.

The poor performance on \ourdataset is due to not only occlusions/crowds/small-scale in static images but also objects' disappearance-reappearance and flickering across the temporal domain. While the heavy occlusions, crowds, and small objects bring challenges to the segmentation of objects in images, the disappearance-reappearance of objects makes it extremely difficult to track an occluded object, increasing the challenge of association.

In a summary, our main contributions are as follows:
\begin{itemize}
\setlength\itemsep{0.1em}
    \item We build a new video object segmentation benchmark dataset termed \ourdataset (co\textbf{M}plex video \textbf{O}bject \textbf{SE}gmentation). \ourdataset focuses on understanding objects in videos under complex environments.
    \item We conduct comprehensive comparison and evaluation of state-of-the-art VOS methods on the \ourdataset dataset under 4 different settings, including mask-initialization semi-supervised, box-initialization semi-supervised, unsupervised, and interactive settings. 
    \item Taking a close look at \ourdataset, we analyze the challenges and potential directions for future video understanding research in complex scenes.
\end{itemize}
\section{Related Work}
\subsection{Video Object Segmentation (VOS)}
Video object segmentation (VOS) aims at segmenting a particular object throughout the entire video clip sequence. According to how to indicate the particular object, there are mainly four different settings, \ie, semi-supervised VOS (or semi-automatic VOS~\cite{wang2021survey} or one-shot VOS), unsupervised VOS (or automatic VOS~\cite{wang2021survey} or zero-shot VOS), interactive VOS, and referring VOS.

\noindent$\bullet$~\textbf{Semi-supervised VOS.} Semi-supervised video object segmentation (or one-shot video object segmentation)~\cite{DBLP:conf/cvpr/CaellesMPLCG17} gives the first frame object mask and target to segment the target object throughout the remaining video frames. Most existing works can be categorized into propagation-based methods~\cite{DBLP:conf/cvpr/PerazziKBSS17,jang2017online,DBLP:conf/cvpr/JampaniGG17,xiao2018monet,hu2018motion,han2018reinforcement,youtube_vos,cheng2018fast,xu2019mhp,chen2020state,huang2020fast,wug2018fast,lin2019agss,zhang2019fast} and matching-based methods~\cite{DBLP:conf/iccv/YoonRKLSK17,cheng2018fast,voigtlaender2019feelvos,wang2019ranet,Duarte_2019_ICCV,Oh_2019_ICCV,zhang2020transductive,yang2020collaborative,Hu_2021_CVPR,duke2021sstvos}. The propagation-based methods utilize the mask of previous frame to guide the mask generation of the current frame, propagating the clues of target object frame by frame. The matching-based methods memorize the embedding of target object and then conduct per-pixel classification to measure the similarity of each pixel's feature to the embedding of target object. 

 Since pixel-wise masks are hard to be obtained, some semi-supervised VOS works propose utilize \textbf{bounding box} as the first-frame clue to indicate the target object~\cite{SimMask, FTMU, lin2021query}. For example, SiamMask~\cite{SimMask} applies a mask prediction branch on fully-convolutional Siamese object tracker to generate binary segmentation masks.

\noindent$\bullet$~\textbf{Interactive VOS.} Interactive video object segmentation is a special form of semi-supervised video object segmentation~\cite{MiVOS,oh2019fast,GIS,MANet,chen2018blazingly,ding2022deep,cheng2018fast,chen2020scribblebox,Yin_2021_CVPR}, it aims at segmenting the target object in a video indicated by user's interaction, \eg, clicks or scribbles. Existing interactive VOS mainly follows a paradigm of interaction-propagation way. Besides the feature encoder that extracts discriminative pixel features, there are other two modules placed on the feature encoder to achieve interactive video object segmentation, \ie, interactive segmentation module that corrects prediction based on user's interaction and mask propagation module that propagates user-corrected masks to other frames.

\noindent$\bullet$~\textbf{Referring VOS.} Referring video object segmentation~\cite{seo2020urvos, liu2022instance, wang2019asymmetric, Ding_2022_CVPR,phraseclick,ye2021referring,ding2021vision} is an emerging setting that involves multi-modal information. It gives a natural language expression to indicate the target object and aims at segmenting the target object throughout the video clips. Existing referring video object segmentation methods can be categorized into two ways: bottom-up methods and top-down methods. Bottom-up methods~\cite{seo2020urvos, VLTPAMI, liu2021cmpc} directly segment the target object at the first frame and propagate the mask to the remaining frames, or conduct image segmentation on each frame independently and then associate these masks. Top-down methods~\cite{liang2021topdown, Botach_2022_CVPR, Wu_2022_CVPR} exhaustively propose all potential tracklets and select the one that is best matched with language expression as output.

\noindent$\bullet$~\textbf{Unsupervised VOS.} It is also known as automatic VOS or zero-shot VOS~\cite{PMOSR,fragkiadaki2015learning,yang2021dystab,RTNet,liu2021f2net,lu2020video,lu2020learning,Tokmakov2019,yang2019anchor,Li_2019_ICCV,wang2019learning2}. Different from the above VOS settings, unsupervised VOS does not require any manual clues to indicate the objects but aims to automatically find the primary objects in a video. Unsupervised VOS can only deal with objects of pre-defined categories. Early methods usually use post-processing steps~\cite{fragkiadaki2015learning}. Then end-to-end training methods become the mainstream in unsupervised VOS, which can be categorized into local content encoding and contextual content encoding. The local content encoding methods~\cite{jain2017fusionseg,cheng2017segflow,Li_2019_ICCV,li2018flow,Tokmakov2019,DBLP:conf/iccv/TokmakovAS17,zhou2020motion} typically employ two parallel networks to extract features from optical flow and RGB image. The contextual content encoding methods~\cite{Lu_2019_CVPR,AGCNN,lu2021segmenting} focus on capturing long-term global information.

\subsection{Related Video Segmentation Tasks} 
There are some other video segmentation tasks that are related to video object segmentation, for example, video instance segmentation, video semantic segmentation, and video panoptic segmentation.

\noindent$\bullet$~\textbf{Video Instance Segmentation (VIS).} Video instance segmentation is extended from image instance segmentation by Yang \etal~\cite{yang2019video}, it simultaneously conducts detection, segmentation, and tracking of instances of predefined categories in videos. Thanks to the large-scale VIS dataset YouTube-VIS~\cite{yang2019video}, a series of learning methods have been developed and greatly advanced the performance of VIS~\cite{TPRPAMI,VMT}. Then, occluded video instance segmentation is proposed by~\cite{OVIS} to study the video instance segmentation under occluded scenes. Similar to~\cite{OVIS}, we study video understanding under complex scenarios like occlusions, but different from~\cite{OVIS}, we focus on video object segmentation (VOS) and our proposed \ourdataset dataset contains more videos and more complex scenes than~\cite{OVIS}, especially in terms of object's disappearance-reappearance.

\noindent$\bullet$~\textbf{Video Semantic Segmentation (VSS).} Driven by the great success in image semantic segmentation~\cite{ding2018context,ding2021interaction,long2015fully,Deeplabv2,BFP,SVCNet} and large-scale video semantic segmentation datasets~\cite{brostow2009semantic,cordts2016cityscapes,miao2021vspw}, video semantic segmentation has drawn lots of attention and achieved significant achievement. Compared to image domain, temporal consistency and model efficiency are the new efforts in the video domain. For example, Sun \etal~\cite{sun2022coarse} propose Coarse-to-Fine Feature Mining (CFFM) to capture both static context and motional context.

\noindent$\bullet$~\textbf{Video Panoptic Segmentation (VPS).} Kim \etal~\cite{kim2020video} introduce panoptic segmentation to the video domain to simultaneously segment and track both the foreground instance objects and background stuff. They build a VPS dataset with 124 videos. Then, Miao \etal~\cite{VIPSeg} build a larger VPS dataset called VIPSeg with 3,536 videos. Existing methods~\cite{Woo_2021_CVPR,Qiao_2021_CVPR} mainly add temporal refinement or cross-frame association modules upon image panoptic segmentation models to enhance temporal conformity and instance tracking performance.

\subsection{Complex Scene Understanding}

Complex scene understanding has become a research focus in the image understanding domain~\cite{shuai2018toward,ding2020semantic,lazarow2020learning,deocclusion,compositional2,repulsion,liu2019feature,orcnn,wang2019bi,chiou2021recovering}. For example, Ke \etal~\cite{bcnet} propose Bilayer Convolutional Network (BCNet) to decouple overlapping objects into occluder and occludee layers. Zhang \etal~\cite{deocclusion} propose a self-supervised approach to conduct de-occlusion by ordering recovery, amodal completion, and content completion. 
On the video domain, however, occlusion understanding is still underexplored with only several multi-object tracking works \cite{mot_attn_occ1,mot_attn_occ2,mot_topo_occ1,mot_topo_occ2}. For example, Chu \etal~\cite{mot_attn_occ1} propose a spatial temporal attention mechanism (STAM) to capture the visible parts of targets and deal with the drift brought by occlusion. Zhu \etal~\cite{mot_attn_occ2} propose dual matching attention networks (DMAN) to deal with the noisy occlusions in multi-object tracking. Li \etal~\cite{TETr} propose to track every thing in the open world by performing class-agnostic association. In this work, we build a Complex Video Object Segmentation dataset to support future work on complex scene understanding in VOS.

\begin{table}[t]
\renewcommand\arraystretch{1.36}
\setlength{\tabcolsep}{1.36pt}
\footnotesize
\centering
\caption{A complete list of object categories and their \#instances in the \ourdataset dataset. The object categories are sorted in descending order of their frequency of occurrence.}
\label{tab:categories}
\setlength{\tabcolsep}{2pt}
\begin{tabular}{|l|r|l|r|l|r|l|r|}
\hline
\rowcolor{mygray} Category & No. &Category & No. &Category & No. &Category & No.\\ \hline\hline
{Person-other}&519&Monkey& 181&Bicycle& 130&{Boat}&73\\
\rowcolor{mygray2}Fish & 343&Dog& 175&{Cyclist}& 116&Lizard& 68 \\
Horse &  272&Boat& 166&Tiger& 108&{Duck}&64\\
\rowcolor{mygray2}Sheep & 264&{Sedan}& 160&Giraffe& 108&{Goose}&54\\
Zebra& 254&{Motorcyclist}& 153&Panda& 106&{Horse-rider}& 46\\
\rowcolor{mygray2}Rabbit&   237&Turtle& 142&{Driver}&90&{Bus}& 29\\
Bird&  226&Cat& 139&Airplane& 89&{Truck}& 15\\
\rowcolor{mygray2}Elephant& 207&Parrot& 139&{Chicken}&86&{Vehicle-other}& 14 \\
Motorcycle& 192&Cow& 139 &Bear& 84&{Poultry-other}&12\\
\hline
\end{tabular}
\end{table}

\begin{table*}[t]
\footnotesize
\centering
\caption{Scale comparison between \ourdataset and existing VOS datasets. ``Annotations'' denotes the number of annotated object masks. ``Duration'' denotes the total duration (in minutes) of the annotated videos. ``mBOR'' denotes the mean of the Bounding-box-Occlusion Rate. ``Disapp. Rate'' represents the frequency of objects that disappear in at least one frame. The newly built \ourdataset has the longest video duration and the largest number of annotations. More importantly, the most notable feature of \ourdataset is that it contains lots of crowds, occlusions, and disappearance-reappearance objects, which provide much more complex scenarios for video object segmentation.}
\label{tab:dataset-cmp}
\renewcommand\arraystretch{1.36}
\setlength{\tabcolsep}{10pt}
\begin{tabular}{r|c|c|c|c|c|c|c|c}
\thickhline
\rowcolor{mygray}Dataset~~~~& Year&Videos & Categories&Objects &Annotations & Duration& mBOR& Disapp. Rate\\
\hline
\hline
{YouTube-Objects}~\cite{DBLP:conf/cvpr/PrestLCSF12} &2012&96 & 10 & 96 & 1,692 & 9.01 &-  &-\\ 
\rowcolor{mygray2}{SegTrack-v2}~\cite{li2013video} &2013&14 & 11 & 24 & 1,475 & 0.59 & 0.12 & 8.3\% \\ 
{FBMS}~\cite{DBLP:journals/pami/OchsMB14} &2014&59 & 16 & 139 & 1,465 & 7.70 & 0.01 & 11.2\% \\
\rowcolor{mygray2}JumpCut~\cite{JumpCut} &2015 &22 & 14 & 22 & 6,331 & 3.52 &  0 &0\%\\

{DAVIS}$_{16}$\cite{davis2016} &2016&50 & - & 50 & 3,440 & 2.88 & - & - \\
\rowcolor{mygray2}{DAVIS}$_{17}$\cite{davis2017} &2017&90&-&205&13,543&5.17& 0.03 & 16.1\% \\
{YouTube-VOS}~\cite{youtube_vos} &2018&4,453& 94& 7,755& 197,272& 334.81& 0.05 & 13.0\% \\
\hline
\rowcolor{mygray2}\textbf{{\ourdataset}} (ours) &\textbf{2023}& \textbf{\numvideo} & \textbf{\numclass} & \textbf{\numobject} & \textbf{\nummask} & \textbf{443.62} &  \textbf{0.23}&\textbf{28.8\%}\\
\hline
\end{tabular}
\end{table*}

\section{\ourdataset Dataset}\label{sec:OVOS_Dataset}
In this section, we introduce the newly built \ourdataset dataset. We first present the video collection and annotation process in \Cref{sec:videoannotation} and then give the dataset statistics and analysis in \Cref{sec:datasetStatistics}. Finally, we give the evaluation metrics in \Cref{sec:Metrics}.

\subsection{Video Collection and Annotation}\label{sec:videoannotation}
\paragraph{Video Collection.} The videos of \ourdataset are from two parts, one is inherited from OVIS~\cite{OVIS}, which is designed for video instance segmentation, and the other is newly captured videos from real-world scenarios. We choose OVIS because it contains lots of heavy occlusions and meets the requirements of complex scenes. For videos inherited from OVIS, since there are many objects that appear for the first time in the non-first frame, these objects cannot be directly used for video object segmentation that requires a first-frame reference. One solution is to discard all the objects that do not appear in the first frame, but this will waste a lot of mask annotation. To adapt OVIS videos to video object segmentation tasks and make full use of them, we cut each of these videos into several videos according to the frame where each object first appears in the videos and then discard videos that do not meet our requirements. 
For the newly captured videos, we first select a set of semantic categories that are common in the real world, including vehicles (\eg, \texttt{bicycle}, \texttt{bus}, \texttt{airplane}, \texttt{boat}), animals (\eg, \texttt{bird}, \texttt{panda}, \texttt{dog}, \texttt{giraffe}), and human beings in different activities (\eg, \texttt{motorcycling}, \texttt{driving}, \texttt{riding}, \texttt{running}), and then collect/film videos containing these categories from the campus, zoo, indoor, city street, etc. A complete list of object categories in the \ourdataset dataset can be seen in \Cref{tab:categories}. 
These categories are common in our life where crowds and occlusions frequently occur. Besides, they are contained in popular large-scale image segmentation datasets like MS-COCO~\cite{coco}, which makes it very easy to use image-pretrained models.

As our primary concern is video object segmentation under complex scenes containing crowded and occluded objects, we have set several rules in order to ensure that crowded and occluded objects are included when collecting/shooting videos:
\begin{itemize}
\setlength\itemsep{0.5em}
\item[R1.] Each video has to contain several objects while those with only a single object are excluded. Especially, videos with crowded objects of similar appearance are highly accredited.

\item[R2.] Occlusions should be present in the video. The videos that do not have any occlusions throughout the entirety of the frames are discarded. Occlusions caused by other moving objects are encouraged.

\item[R3.] Great emphasis should be placed on scenarios where objects disappear and then reappear due to occlusions or crowding.

\item[R4.] The target objects should be of a variety of scales and types, \eg, small scale or large scale, conspicuous or inconspicuous.

\item[R5.] Objects in the video must show sufficient motions. Videos with completely still objects or very little motions are culled.
\end{itemize}

Aside from the above rules, in order to guarantee the video quality, we require video resolution to be $1920\times1080$ and video length to be 5 to 60 seconds in general in order to make sure that the video is of high quality.

\paragraph{Video Annotation.} Having all videos for \ourdataset been collected, our research team looks through them and figure out a set of targets-of-interest for each of the videos. Then we slightly clip the start and the end of videos, to reduce the number of less motional or simple frames in the video. Next we annotate the first-frame mask of the target objects, as VOS input. Following this, the videos are sent to the annotation team along with the first-frame masks for annotation of the subsequent video frames.  

Using the given first-frame mask as a reference, the annotation team is required to identify the target object in the given first-frame mask, then track and annotate the segmentation mask of the target object in all frames following the first frame. The process of annotating videos has been made easier with the help of an interactive annotation tool that we developed. The annotation tool automatically loads videos and all target objects. Annotators use the tool to load and preview videos and first-frame masks, annotate and visualize the segmentation masks in the subsequent frames, and save them. The annotation tool also has a built-in interactive object segmentation network XMem~\cite{XMem}, to assist annotations in producing high-quality masks.
To ensure the annotation quality under complex scenes, the annotators are required to clearly track the object that disappears and reappears due to heavy occlusions and crowd. For frames in which the target object is disappeared or is fully occluded, annotators also need to confirm that the output masks of such frames shall be blank. It is a requirement that all of our videos be annotated every five frames at the very least. For the purpose of testing the frame-rate robustness of the models, some videos are annotated every frame. 

Following the annotation, the videos are sent to the verification team for verification of annotation quality in all the areas, especially annotations of videos with occlusions, a crowd, or disappeared-reappeared target objects.

\subsection{Dataset Statistics}\label{sec:datasetStatistics}

In~\Cref{tab:dataset-cmp}, we analyze the data statistics of the new \ourdataset dataset using previous datasets for the video object segmentation as a reference for the analysis, including JumpCut \cite{JumpCut}, SegTrack-v2 \cite{li2013video}, {YouTube-Objects} \cite{DBLP:conf/cvpr/PrestLCSF12}, {FBMS} \cite{DBLP:journals/pami/OchsMB14}, {DAVIS} \cite{davis2016,davis2017}, and {YouTube-VOS} \cite{youtube_vos}. As shown in \Cref{tab:dataset-cmp}, there are \numvideo videos and \nummask mask annotations for \numobject objects contained in the proposed \ourdataset dataset. Comparing with previous largest VOS dataset YouTube-VOS~\cite{youtube_vos}, MOSE has much more mask annotations, 431k \vs 197k. Among the 8 video object segmentation datasets that we analyzed in~\Cref{tab:dataset-cmp}, \ourdataset has the longest video duration (433.62 minutes), which is about 100 minutes longer than the second longest dataset YouTube-VOS (334.81 minutes) and is hundreds of times longer than the other seven datasets. We include more long videos in the \ourdataset than previous VOS datasets to ensure that we have a variety of occlusion scenarios, motion scenarios, and object disappearance-reappearance in our \ourdataset dataset.

\begin{figure}[t]
\centering
         \includegraphics[width=\linewidth]{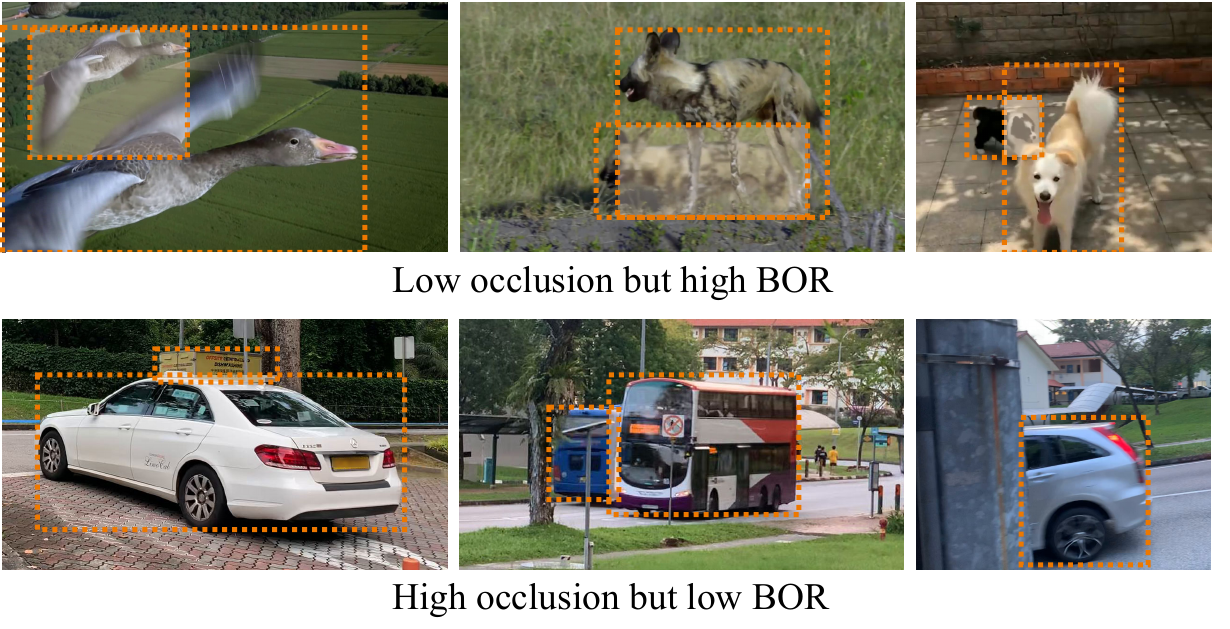}
         \caption{Failure cases of the BOR indicator. It can be seen from the first row of samples that they have high BOR values, but there is less or no occlusion present. Samples in the second row have very small BOR values, but there are severe occlusions in the samples.}
      \label{fig:borfaliure}
\end{figure}

\paragraph{Disappearance and Occlusion Analysis.} Previous occluded video datasets OVIS~\cite{OVIS} defines a Bounding-box Occlusion Rate (BOR) to reflect the occlusion degree. The BOR is calculated by the Intersection-over-Union of bounding boxes in the videos.
We provide the mean of BOR of all frames in \Cref{tab:dataset-cmp}. The \ourdataset has the largest value of mBOR, indicating the more frequency of occlusions in \ourdataset than previous datasets. However, we find that BOR can only roughly reflect the occlusion. It does not well reveal the degree of occlusion in the \ourdataset dataset and may be prone to making some mistakes in this regard. As shown in the first row of~\Cref{fig:borfaliure}, high BOR is observed in the three images with low occlusion or even no occlusion. In contrast, in the images of the second row, where there are heavily occluded objects, the BOR is low and even 0.

Therefore, besides counting BOR, we further calculate the number of disappeared objects that disappear in at least one frame of the video, followed by the disappearance rate that reflects the frequency of disappearances, which illustrates how often the disappearance of objects occurs due to complex scenarios in a dataset.
The number of objects with disappearance in \ourdataset is 1,553, which is much higher than the previous datasets. As shown in \Cref{tab:dataset-cmp}, the disappearance rate (Disapp. Rate) of \ourdataset is the highest 28.8\%, reflecting that the disappearance and occlusions are frequent and severe in \ourdataset.

\vspace{3mm}
\subsection{Evaluation Metrics}\label{sec:Metrics}
In accordance with the previous method~\cite{davis2016,davis2017}, we compute the region similarity \(\mathcal{J}\) and the contour accuracy \(\mathcal{F}\) as evaluation metrics. Given segmentation mask predictions ${\hat{M}}\!\in\!\{0,1\}^{H\times W}$ and the ground-truth masks ${M}\!\in\!\{0,1\}^{H\times W}$, region similarity \(\mathcal{J}\) is obtained by calculating the Intersection-over-Union (IoU) of ${\hat{M}}$ and ${{M}}$,
\begin{equation}\small
\mathcal{J} = \frac{{\hat{M} \cap  M}}{{\hat{M}\cup M}}.
\vspace{-2pt}
\end{equation}
Then, the average region similarity \(\mathcal{J}_{mean}\) over all objects is calculated as the final region similarity result. We use \(\mathcal{J}\) to represent \(\mathcal{J}_{mean}\) for the sake of brevity. To measure the contour quality of ${\hat{M}}$, contour recall $\text{R}_c$ and precision $\text{P}_c$ are calculated via bipartite graph matching~\cite{martin2004learning}. Then, the contour accuracy \(\mathcal{F}\) is the harmonic mean of the contour recall $\text{R}_c$ and precision $\text{P}_c$, \ie, 
\begin{equation}\small
	\mathcal{F} = \frac{{2 \text{P}_c \text{R}_c}}{{\text{P}_c + \text{R}_c}},
	\label{f-compute}
\end{equation}
which represents how closely the contours of predicted masks resemble the contours of ground-truth masks. The average contour accuracy \(\mathcal{F}_{mean}\) over all objects is calculated as the final contour accuracy result. We use \(\mathcal{F}\) to represent \(\mathcal{F}_{mean}\) for the sake of brevity. Then, $\mathcal{J}\&\mathcal{F}=(\mathcal{J}+\mathcal{F})/2$ is used to measure the overall performance.
\section{Experiments}
Herein we conduct experiments and benchmarks on four different Video Object Segmentation (VOS) settings, including semi-supervised (or one-shot) VOS with mask-initialization, semi-supervised VOS with box-initialization, unsupervised (or zero-shot) VOS, and interactive VOS, to comprehensively analyze the newly built \ourdataset dataset.

\vspace{-3mm}
\paragraph{Implementation Details.}
The proposed \ourdataset dataset is consistent with the YouTube-VOS~\cite{youtube_vos} format. We replace the training dataset of previous methods from YouTube-VOS with our \ourdataset and strictly follow their training settings on YouTube-VOS~\cite{youtube_vos}. We follow DAVIS~\cite{davis2016,davis2017} to evaluate the performance and report the results of \(\mathcal{J}_{mean}\), \(\mathcal{F}_{mean}\), and \(\mathcal{J}\&\mathcal{F}\) on the validation set of \ourdataset. During the training of these VOS methods, we follow their way of using image pre-trained models but do not use any additional video datasets for pretraining.

\vspace{-3mm}
\paragraph{Settings.}
There are \numvideo videos in the whole \ourdataset dataset. These videos are split into 1,507 training videos, 311 validation videos, and 331 testing videos, for model training, daily evaluation, and competition period evaluation, respectively. Each of the videos gives a first-frame mask or bounding box as the reference of the target object for the semi-supervised (one-shot) VOS setting, and first-frame scribbles for the interactive VOS setting. 

\begin{table}[t]
\centering
\renewcommand\arraystretch{1.36}
\setlength{\tabcolsep}{2.4pt}
\footnotesize
\caption{Comparisons of state-of-the-art semi-supervised methods on the validation set. ``$\mathcal{J}$" and ``$\mathcal{F}$" denote the mean of region similarity and the mean of contour accuracy. $\mathcal{J}\&\mathcal{F}$ denotes the mean of $\mathcal{J}$ and $\mathcal{F}$. BL30K~\cite{MiVOS} is not added during the training stage to make a fair comparison.}
\label{tab:svos}
\begin{tabular}{l|c|ccc|c|c}
\hline
\rowcolor{mygray3}       & &\multicolumn{3}{c|}{\textbf{\ourdataset (ours)}} & DAVIS$_{17}$ & YT-VOS$_{18}$ \\
\rowcolor{mygray3}\multirow{-2}{*}{Method}& \multirow{-2}{*}{Pub.}&$\mathcal{J}$ & $\mathcal{F}$ &$\mathcal{J}\&\mathcal{F}$ & $\mathcal{J}\&\mathcal{F}$ & $\mathcal{G}$\\ \hline
AOT~\cite{AOT}&NeurIPS'21&53.1&61.3&57.2&84.9&84.1\\
STCN~\cite{STCN} &NeurIPS'21&46.6&55.0&50.8&85.4&83.0\\
RDE~\cite{RDE}&CVPR'22&44.6&52.9&48.8 &84.2&-\\
SWEM~\cite{SWEM}&CVPR'22&46.8&54.9&50.9 &84.3&82.8\\
XMem~\cite{XMem} &ECCV'22&53.3&62.0&57.6&86.2& 85.7\\
DeAOT~\cite{DeAOT}&NeurIPS'22&55.1&63.8&59.4&85.2&86.0\\
\hline
\end{tabular}
\end{table}
\subsection{Semi-supervised Video Object Segmentation}
Semi-supervised video object segmentation, or one-shot video object segmentation, gives either the first frame mask or bounding box of the target object as a clue and reference. We train and evaluate the very recent 6 mask-initialization semi-supervised VOS methods and 2 box-initialization semi-supervised VOS methods built upon ResNet-50~\cite{resnet} on the \ourdataset dataset, as shown in \Cref{tab:svos} and \Cref{tab:box-svos}, respectively. It is our hope that the experiments will provide baselines for future semi-supervised VOS algorithms to be developed.

\vspace{1mm}
\textbf{Mask-initialization.}~This setting is a classic and currently the most popular topic for video object segmentation. A lot of excellent deep-learning-based works have been developed for this setting in the past decade and greatly improve the video object segmentation performance to a saturation level. For example, the most recent state-of-the-art method DeAOT~\cite{DeAOT} achieves \textbf{85.2\% $\mathcal{J}\&\mathcal{F}$} on the DAVIS 2017~\cite{davis2017}, \textbf{86.0\% $\mathcal{G}$} on YouTube-VOS~\cite{youtube_vos}, and \textbf{92.3\% $\mathcal{J}\&\mathcal{F}$} on DAVIS 2016~\cite{davis2016}, which are excellent performance and almost reach the ground truth. To analyze the newly built \ourdataset dataset and test the performance of existing methods in complex scenes, we train and evaluate these methods on \ourdataset. We report the results of 6 recent state-of-the-art methods on the validation set of \ourdataset, including AOT~\cite{AOT}, STCN~\cite{STCN}, RDE~\cite{RDE}, SWEM~\cite{SWEM}, XMem~\cite{XMem}, and DeAOT~\cite{DeAOT}, as shown in \Cref{tab:svos}. Current state-of-the-art methods only achieve the performance from \textbf{48.8\% $\mathcal{J}\&\mathcal{F}$} to \textbf{59.4\% $\mathcal{J}\&\mathcal{F}$} on the validation set of the newly proposed \ourdataset dataset, while their results on DAVIS 2017~\cite{davis2017} and YouTube-VOS~\cite{youtube_vos} are usually above \textbf{80\% $\mathcal{J}\&\mathcal{F}$} or $\mathcal{G}$, some of them almost reach \textbf{90\% $\mathcal{J}\&\mathcal{F}$} or $\mathcal{G}$ on DAVIS 2017 and YouTube-VOS. The results reveal that although we have achieved excellent VOS performance on previous benchmarks, there are unresolved challenges under complex scenes and more efforts are desired to explore these challenges in the future. Discussion about the \ourdataset dataset and some potential future directions are provided in \Cref{sec:Discussion}.

\vspace{1mm}
\textbf{Box-initialization.} As shown in \Cref{tab:box-svos}, two semi-supervised (one-shot) VOS methods with a bounding box as the first-frame reference are trained on \ourdataset and evaluated on the validation set of \ourdataset, including SimMask~\cite{SimMask} and FTMU~\cite{FTMU}. Current state-of-the-art methods achieve only \textbf{22.0\%} and \textbf{23.8\%} $\mathcal{J}\&\mathcal{F}$ on the validation set of the newly proposed \ourdataset dataset, while their results on DAVIS 2017~\cite{davis2017} are already \textbf{70+\%} $\mathcal{J}\&\mathcal{F}$. 

After careful analysis, we believe that one of the reasons for such a significant drop in performance is that there are many heavy occlusions, small-scale objects, and crowded scenarios in the videos of \ourdataset dataset, which make the weakly supervised segmentation of the videos much more difficult. There is a possibility that, even within the target bounding box region, the target object may not the most salient one due to heavy occlusions and crowd. The occlusion of an object can cause it to be broken up into several pieces that are not adjacent to each other, which greatly increases the difficulty of segmenting an object using a box-driven approach.

\begin{table}[t]
\centering
\renewcommand\arraystretch{1.2}
\setlength{\tabcolsep}{3pt}
\footnotesize
\caption{Comparisons of state-of-the-art box-initialization semi-supervised methods on the validation set. ``$\mathcal{J}$" and ``$\mathcal{F}$" denote the mean of region similarity and the mean of contour accuracy. $\mathcal{J}\&\mathcal{F}$ denotes the mean of $\mathcal{J}$ and $\mathcal{F}$.}
\label{tab:box-svos}
\begin{tabular}{l|c|ccc|c|c}
\hline
\rowcolor{mygray3}       & &\multicolumn{3}{c|}{\textbf{\ourdataset (ours)}} & DAVIS$_{17}$ & YT-VOS$_{18}$ \\
\rowcolor{mygray3}\multirow{-2}{*}{Method}& \multirow{-2}{*}{Pub.}&$\mathcal{J}$ & $\mathcal{F}$ &$\mathcal{J}\&\mathcal{F}$ & $\mathcal{J}\&\mathcal{F}$ & $\mathcal{G}$\\ \hline
SiamMask~\cite{SimMask}  &CVPR'19&17.3&26.7&22.0&56.4& 52.8   \\ 
FTMU~\cite{FTMU}  &CVPR'20&19.1&28.5&23.8&70.6& - \\ \hline
\end{tabular}
\end{table}

\subsection{Unsupervised Video Object Segmentation}
Unsupervised video object segmentation, or zero-shot video object segmentation, does not require any manual clues (\eg, mask or bounding box) as a reference to indicate the objects but aims to find the primary objects in a video automatically. The mainstream of zero-shot VOS methods focuses on addressing single-object VOS. However, \ourdataset is a multi-object VOS dataset like DAVIS 2017~\cite{davis2017}, thus we only benchmark multi-object zero-shot VOS methods on \ourdataset, \eg, STEm-Seg~\cite{athar2020stem}, AGNN~\cite{AGCNN}, and RVOS~\cite{ventura2019rvos}. The results are shown in \Cref{tab:uvos}. In this setting, only videos with exhaustive first-frame annotations are used. Existing methods rely on off-the-shelf image-trained instance segmentation methods, which can well detect/segment objects in the static image thanks to complex scene learning in the image domain. The performance drop is mainly due to temporal challenges.

\begin{table}[t]
\centering
\renewcommand\arraystretch{1.2}
\setlength{\tabcolsep}{6pt}
\footnotesize
\caption{Comparisons of state-of-the-art multi-object zero-shot VOS methods on the validation set. ``$\mathcal{J}$" and ``$\mathcal{F}$" denote the mean of region similarity and the mean of contour accuracy. $\mathcal{J}\&\mathcal{F}$ denotes the mean of $\mathcal{J}$ and $\mathcal{F}$.}
\label{tab:uvos}
\begin{tabular}{l|c|ccc|c}
\hline
\rowcolor{mygray3}       & &\multicolumn{3}{c|}{\textbf{\ourdataset (ours)}} & DAVIS$_{17}$ \\
\rowcolor{mygray3}\multirow{-2}{*}{Method}& \multirow{-2}{*}{Pub.}&$\mathcal{J}$ & $\mathcal{F}$ &$\mathcal{J}\&\mathcal{F}$ & $\mathcal{J}\&\mathcal{F}$ \\ \hline
RVOS~\cite{ventura2019rvos}&CVPR'19&24.1&36.9&30.5&43.7\\
AGNN~\cite{AGCNN}&ICCV'19&38.6&48.8&43.7&61.1\\
STEm-Seg~\cite{athar2020stem}&ECCV'20&43.3&50.5&46.9&64.7 \\
\hline
\end{tabular}
\end{table}

\begin{table}[t]
\centering
\renewcommand\arraystretch{1.2}
\setlength{\tabcolsep}{9pt}
\footnotesize
\caption{Comparisons of state-of-the-art interactive VOS methods on the validation set. $\mathcal{J}\&\mathcal{F}$@60s denotes the $\mathcal{J}\&\mathcal{F}$ performance reached by methods within 60 seconds interactions.}
\label{tab:ivos}
\begin{tabular}{l|c|c|c}
\hline
\rowcolor{mygray3}       & &{\textbf{\ourdataset (ours)}} & DAVIS$_{17}$ \\
\rowcolor{mygray3}\multirow{-2}{*}{Method}& \multirow{-2}{*}{Pub.}&$\mathcal{J}\&\mathcal{F}$@60s & $\mathcal{J}\&\mathcal{F}$@60s \\ \hline
IPNet~\cite{oh2019fast}&CVPR'19& 41.2 &78.7\\
STM~\cite{stm}&ICCV'19& 45.3 &84.8\\
MANet~\cite{MANet}&CVPR'20& 43.6 &79.5\\
ATNet~\cite{heo2020interactive}&ECCV'20& 44.5 &82.7\\
GIS~\cite{GIS}&CVPR'21& 50.2 &86.6\\
MiVOS~\cite{MiVOS}&CVPR'21& 51.4 &88.5\\
STCN~\cite{STCN} &NeurIPS'21& 56.8 & 88.8\\
\hline
\end{tabular}
\end{table}
\subsection{Interactive Video Object Segmentation}
Following the interactive track of DAVIS 2019 Challenge~\cite{davis2019}, we provide initial scribbles for the target object in a specific video sequence, which are used as the first interaction. The interactive video object segmentation methods are required to predict the segmentation mask for the whole video based on the first interaction. Then, by comparing predicted masks and ground truth masks across the video, corrective scribbles on the worst frame are further provided for the methods to refine the video segmentation prediction. The above step is allowed to be repeated up to 8 times with a time limitation of 30s for each object. We report the metric of $\mathcal{J}\&\mathcal{F}$@60s to encourage the methods to have a good balance between speed and accuracy. As shown in \Cref{tab:ivos}, seven recent interactive video object segmentation methods are evaluated on the validation set of \ourdataset.
\section{Discussion and Future Directions}\label{sec:Discussion}
Herein we discuss the challenges of the proposed \ourdataset dataset brought by complex scenes and provide some potential future directions based on the experimental results analysis of the existing methods on \ourdataset.

\vspace{1mm}
\noindent$\bullet$~\textbf{Stronger Association to Track Reappearing Objects.} It is necessary to develop stronger association/re-identification algorithms for VOS methods in order to be able to track objects that disappear and then reappear. Especially, the most interesting thing is that we have noticed that a number of disappeared-then-reappeared objects have been reappearing with a different appearance from the time they disappeared, \ie, appearance-changing objects. For example, the target player in \Cref{Fig:teaser} shows his back before disappearing while showing the front at reappearing. There is a great deal of difficulty in tracking with such an appearance change.

\vspace{1mm}
\noindent$\bullet$~\textbf{Video Object Segmentation of Occluded Objects.} The frequent occlusions in the videos of \ourdataset provide data support to the research of occlusion video understanding. Tracking and segmenting an object with heavy occlusion have rarely been studied since the target objects in existing datasets are usually salient and dominant. In real-world scenarios, isolated objects rarely appear while occluded scenes occur frequently. Our human beings can well capture those occluded objects, thanks to our contextual association and reasoning ability. With the introduction of an understanding of occlusion into the process of video object segmentation, the VOS methods will become more practical for use in real-world applications. Especially, the occlusions make the box-initialization semi-supervised VOS setting more challenging in terms of segmenting an occluded object by bounding box.

\vspace{1mm}
\noindent$\bullet$~\textbf{Attention on Small \& Inconspicuous Objects.} Although the detection and segmentation of small objects is a hot topic in the image domain, tracking and segmenting small and inconspicuous objects in the video object segmentation domain is still to be developed. As a matter of fact, most of the existing video object segmentation methods mainly focus on tracking large and salient objects. The lack of sufficient attention to small-scale objects makes the video object segmentation methods less pronounced in practical applications that may involve target objects of varying sizes and types. There are many objects in the proposed \ourdataset dataset that are on a small scale and are inconspicuous in the videos, which provides a greater opportunity for the research of tracking and segmenting objects that are small and inconspicuous in realistic scenarios.

\vspace{1mm}
\noindent$\bullet$~\textbf{Tracking Objects in Crowd.} One of the most notable features of the \ourdataset is crowded scenarios, which are common in real-world applications. There are many videos in the \ourdataset dataset, which contain crowd objects, such as flocks of sheep moving together, groups of cyclists racing, pedestrians moving on a crowded street, and \etc. A scenario like this presents challenges when it comes to segmenting and tracking one object among a crowd of objects that share a similar appearance and motion to the target object with the crowd as a whole. In the image/frame domain, it is desired for video object segmentation algorithms to enhance the identification ability to distinguish between different objects with a similar appearance. What's more, in the temporal domain, it is challenging to perform the object association between the two frames with the crowd consisting of similar-looking objects.

\vspace{1mm}
\noindent$\bullet$~\textbf{Long-Term Video Segmentation}: In terms of practical applications like movie editing and surveillance monitoring, long-term video understanding is much more practical than short-term video understanding. For example, the average length of videos on YouTube is around twelve minutes, which is much longer than the average length of the existing video object segmentation dataset. Although existing VOS methods provide excellent performance, they typically require a lot of computing resources, \eg, GPU memory, for storing the object features of previous frames. Due to the large computation cost, most existing VOS methods cannot well handle long videos. The average video length of the \ourdataset dataset is longer than existing VOS datasets, bringing more challenges and opportunities in dealing with long-term videos. It is a good and practical direction to design VOS algorithms that can deal with long videos with low computation costs while achieving high-quality segmentation results.

\section{Conclusion}
We build a large-scale video object segmentation dataset named \ourdataset to revisit and support the research of video object segmentation under complex scenes. There are \numvideo high-resolution videos in \ourdataset with \nummask high-quality object masks for \numobject objects from \numclass categories. The videos in \ourdataset are commonly long enough to ensure diverse and sufficient occlusion, motion, and disappearance-reappearance scenarios. Based on the proposed \ourdataset dataset, we benchmark existing VOS methods and conduct a comprehensive comparison. We train and evaluate six methods under the semi-supervised (one-shot) setting with a mask as the first-frame reference, two methods with a bounding box as the first-frame reference, three methods under the unsupervised (zero-shot) setting, and seven methods under the interactive setting. After evaluating existing VOS methods on \ourdataset dataset and comprehensively analyzing the results, some challenges and potential directions for future VOS research are concluded. We find that we are still at a nascent stage of segmenting and tracking objects in complex scenes where crowds, disappearing, occlusions, and inconspicuous/small objects occur frequently.

{\footnotesize
\bibliographystyle{unsrt}
\bibliography{egbib}
}

\end{document}